# I J M I

# AN IMAGE BASED TECHNIQUE FOR ENHANCEMENT OF UNDERWATER IMAGES

## PRABHAKAR C.J.[1*], PRAVEEN KUMAR P.U.[1]

[1]Department of P.G. Studies and Research in Computer Science
Kuvempu University, Shankaraghatta-577451, Karnataka, India.
*Corresponding Author: Email- psajjan@yahoo.com

**Abstract -** The underwater images usually suffers from non-uniform lighting, low contrast, blur and diminished colors. In this paper, we proposed an image based preprocessing technique to enhance the quality of the underwater images. The proposed technique comprises a combination of four filters such as homomorphic filtering, wavelet denoising, bilateral filter and contrast equalization. These filters are applied sequentially on degraded underwater images. The literature survey reveals that image based preprocessing algorithms uses standard filter techniques with various combinations. For smoothing the image, the image based preprocessing algorithms uses the anisotropic filter. The main drawback of the anisotropic filter is that iterative in nature and computation time is high compared to bilateral filter. In the proposed technique, in addition to other three filters, we employ a bilateral filter for smoothing the image. The experimentation is carried out in two stages. In the first stage, we have conducted various experiments on captured images and estimated optimal parameters for bilateral filter. Similarly, optimal filter bank and optimal wavelet shrinkage function are estimated for wavelet denoising. In the second stage, we conducted the experiments using estimated optimal parameters, optimal filter bank and optimal wavelet shrinkage function for evaluating the proposed technique. We evaluated the technique using quantitative based criteria such as a gradient magnitude histogram and Peak Signal to Noise Ratio (PSNR). Further, the results are qualitatively evaluated based on edge detection results. The proposed technique enhances the quality of the underwater images and can be employed prior to apply computer vision techniques.

## 1. INTRODUCTION

Underwater vision is one of the scientific fields of investigation for researchers. Autonomous Underwater Vehicles (AUV) and Remotely Operated Vehicles (ROV) are usually employed to capture the data such as underwater mines, shipwrecks, coral reefs, pipelines and telecommunication cables from the underwater environment. Underwater images are essentially characterized by their poor visibility because light is exponentially attenuated as it travels in the water, and the scenes result poorly contrasted and hazy. Light attenuation limits the visibility distance at about twenty meters in clear water and five meters or less in turbid water. The light attenuation process is caused by absorption and scattering, which influence the overall performance of underwater imaging systems. Forward scattering generally leads to blur of the image features. On the other hand, backscattering generally limits the contrast of the images, generating a characteristic veil that superimposes itself on the image and hides the scene. Absorption and scattering effects are not only due to the water itself but also due to the components such as a dissolved organic matter. The visibility range can be increased with artificial illumination of light on the object, but it produces non-uniform of light on the surface of the object and producing a bright spot in the center of the image with poorly illuminated area surrounding it. The amount of light is reduced when we go deeper, colors drop off depending on their wavelengths. The blue color travels across the longest in the water due to its shortest wavelength. Underwater image suffers from limited range visibility, low contrast, non-uniform lighting, blurring, bright artifacts, color diminished and noise.

The research on underwater image processing can be addressed from two different points of view such as an image restoration or an image enhancement method [1, 16-18]. The image restoration aims to recover a degraded image using a model of the degradation and of the original image formation; it is essentially an inverse problem. These methods are rigorous, but they require many model parameters like attenuation and diffusion coefficients that characterize the water turbidity and can be extremely variable. Whereas image enhancement uses qualitative subjective criteria to produce a more visually pleasing image and they do not rely on any physical model for the image formation. These kinds of approaches are usually simpler and faster than deconvolution methods. Recently, many researchers have developed preprocessing techniques for underwater images using image enhancement methods. Bazeille et al. [3] propose an algorithm to pre-process underwater images. It reduces underwater perturbations and improves image quality. The algorithm is automatic and requires no parameter adjustment. The method was used as a preliminary step of edge detection. The robustness of the method was analyzed using gradient magnitude histograms and also the criterion used by Arnold-Bos et al. [2] was applied. This criterion assumes that well-contrasted and noise-free images have a distribution of the gradient magnitude histogram close to exponential, and it attributes a mark from zero to one.

Chambah et al. [4] proposed a color correction method based on the Automatic Color Equalization (ACE) model, an unsupervised color equalization algorithm developed by Rizzi et al. [5]. ACE is a perceptual approach inspired by some adaptation mechanisms of the human vision system, in particular, lightness constancy and color constancy. ACE was applied on videos taken in an aquatic environment that present a strong and non-uniform color cast due to the depth of the water and the artificial illumination. Images were taken from the tanks of an aquarium. Iqbal et al. [6] presented an underwater image enhancement method using an integrated color model. They proposed an approach based on a slide stretching: first, contrast stretching of RGB algorithm is used to equalize the color contrast in the images. Second, saturation and intensity stretching of HSI is applied to increase the true color and solve the problem of lighting. The blue color component in the image is controlled by the saturation and intensity to create the range from pale blue to deep blue. The contrast ratio is therefore controlled by decreasing or increasing its value.

Arnold-Bos et al. [2] presented a complete pre-processing framework for underwater images. They investigated the possibility of addressing the whole range of noises present in underwater images by a combination of deconvolution and enhancement methods. First, a contrast equalization system is proposed to reject backscattering, attenuation and lighting inequalities. If $I(i, j)$ is the original image and $I_{LP}(i, j)$ its low-pass version, a contrast-equalized version of $I$ is $I_{eq} = I/I_{LP}$. The additional use of adaptive smoothing helps to address the remaining sources of noise, which is corresponding to sensor noise, floating particles and miscellaneous quantification errors. This method applies local contrast equalization method as a first step in order to deal with non-uniform lighting caused by backscattering. Generally, contrast equalization will raise the noise level in poorly contrasted areas of the original image. Signal to noise ratio would remain constant after equalization; but the fixed color quantization step induces strong errors in dark zones. Compared to the local contrast equalization method, the homomorphic filtering (which adopts the illumination-reflectance model) has a slightly more important effect on noise in dark zones.

In this paper, we propose a preprocessing technique, which consists of sequentially applying filters such as homomorphic filtering, wavelet denoising, bilateral filtering and contrast stretching. First, we apply homomorphic filter to correct non-uniform illumination of light. Homomorphic filter simultaneously normalizes the brightness across an image and increases contrast. The homomorphic filtering performs in the frequency domain and it adopts the illumination and reflectance model. Wavelet based image denoising techniques are necessary to remove random additive Gaussian noise while retaining as much as possible the important image features. The main objective of these types of random noise removal is to suppress the noise while preserving the original image details. We use the bilateral filter to smooth the image while preserving edges and enhance them. Finally, we apply contrast stretching for normalizing the RGB values. The remaining sections of the paper are organized as follows: section 2 describes proposed technique in detail. The experimental results are presented in the section 3. Finally, the section 4 concludes the paper.

## 2. A PREPROCESSING ALGORITHM
In this section, we present filters, which are adopted in the proposed technique. These filters are employed sequentially on degraded images.

### Homomorphic Filtering
Homomorphic filtering is used to correct non-uniform illumination and to enhance contrasts in the image. It's a frequency filtering, preferred to others' techniques because it corrects non-uniform lighting and sharpens the image features at the same time. We consider that image is a function of the product of the illumination and the reflectance as shown below.

$$f(x, y) = i(x, y) \cdot r(x, y), \qquad (1)$$

where $f(x, y)$ is the image sensed by the camera, $i(x, y)$ the illumination multiplicative factor, and $r(x, y)$ the reflectance function. If we take into account this model, we assume that the illumination factor changes slowly through the view field; therefore it represents low frequencies in the Fourier transform of the image. On the contrary reflectance is associated with high frequency components. By multiplying these components by a high-pass filter we can then suppress the low frequencies i.e., the non-uniform illumination in the image. The algorithm can be decomposed as follows:

Separation of the illumination and reflectance components by taking the logarithm of the image. The logarithm converts the multiplicative into an additive one.

$$g(x, y) = \ln(f(x, y)) = \ln(i(x, y) \cdot r(x, y))$$

$$g(x, y) = \ln(i(x, y)) + \ln(r(x, y)). \qquad (2)$$

Computation of the Fourier transform of the log-image

$$G(w_x, w_y) = I(w_x, w_y) + R(w_x, w_y). \qquad (3)$$

High-pass filtering is applied to the Fourier transform decreases the contribution of low frequencies (illumination) and also amplifies the contribution of mid and high frequencies (reflectance), sharpening the image features of the objects in the image

$$S(w_x, w_y) = H(w_x, w_y) \cdot I(w_x, w_y) \\ + H(w_x, w_y) \cdot R(w_x, w_y), \quad (4)$$

with,

$$H(w_x, w_y) = (r_H - r_L) \cdot (1 - \exp(-(\frac{w_x^2 + w_y^2}{2\delta_w^2}))) + r_L. \ (5)$$

where, $r_H = 2.5$ and $r_L = 0.5$ are the maximum and minimum coefficients values and $\delta_w$ a factor which controls the cut-off frequency. These parameters are selected empirically. Computations of the inverse Fourier transform to come back in the spatial domain and then taking the exponent to obtain the filtered image.

## Wavelet Denoising

Thresholding is a simple non-linear technique, which operates on one wavelet coefficient at a time. In its most basic form, each coefficient is thresholded by comparing against threshold, if the coefficient is smaller than threshold, set to zero; otherwise it is kept or modified. Replacing the small noisy coefficients by zero and inverse wavelet transform on the result may lead to reconstruction with the essential signal characteristics and with the less noise. A simple denoising algorithm that uses the wavelet transform consist of the following three steps, (1) calculate the wavelet transform of the noisy signal (2) Modify the noisy detail wavelet coefficients according to some rule (3) compute the inverse transform using the modified coefficients.

Let us consider a signal $\{f_{ij}, i, j = 1, ..., N\}$, where $N$ is some integer power of 2. It has been corrupted by additive noise and one observes

$$g_{ij} = f_{ij} + \varepsilon_{ij}, \quad i, j = 1, ..., N \quad (6)$$

where $\varepsilon_{ij}$ are independent and identically distributed (iid) zero mean, white Gaussian noise with standard deviation $\sigma$ i.e. $N(0, \sigma^2)$ and independent of $f_{ij}$. From the noisy signal $g_{ij}$ we want to find an approximation $\hat{f}_{ij}$. The goal is to remove the noise, or denoise $g(i, j)$, and to obtain an estimate $\hat{f}_{ij}$ and $f_{ij}$ which minimizes the mean squared error (MSE),

$$MSE(\hat{f}) = \frac{1}{N^2} \sum_{i,j=1}^{N} (\hat{f}_{ij} - f_{ij})^2. \quad (7)$$

Let $\mathbf{g} = \{g_{ij}\}_{i,j}$, $\mathbf{f} = \{f_{ij}\}_{i,j}$, and $\mathbf{\varepsilon} = \{\varepsilon_{ij}\}_{i,j}$; that is, the boldfaced letters will denote the matrix representation of the signals under consideration. Let $D = w_{\mathbf{g}}$, $C = w_{\mathbf{f}}$, and $Z = w_{\mathbf{\varepsilon}}$ denote the matrix of wavelet coefficients $\mathbf{g}, \mathbf{f}, \mathbf{\varepsilon}$ respectively. Where, $w$ is the two-dimensional dyadic orthogonal wavelet transform operator. It is convenient to label the subbands of the wavelet transform. The subbands, $HH_k$, $HL_k$, $LH_k$ are called the details, where

$k = 1, ..., J$ is the scale, with $J$ being the largest (or coarsest) scale in the decomposition and a subband at scale $k$ has size $N / 2^k \times N / 2^k$. The subband $LL_j$ is the low resolution residual and is typically chosen large enough such that $N / 2^j \leq N$, $N / 2^j > 1$. The wavelet based denoising method filters each coefficient $g_{ij}$ from the detail subbands with a threshold function to obtain $\hat{f}_{ij}$. The denoised estimate is then $\hat{g} = w^{-1}\hat{f}$, where $w^{-1}$ is the inverse wavelet transform.

Wavelet transform of noisy signal should be taken first and then thresholding function is applied on it. Finally the output should be undergone inverse wavelet transformation to obtain the estimate $\hat{f}$. There are two thresholding functions frequently used, i.e. a hard threshold and soft threshold. The hard-thresholding function keeps the input if it is larger than the threshold; otherwise, it is set to zero. It is described as:

$$\eta_1(w) = wI(|w| > T), \quad (8)$$

where $w$ is a wavelet coefficient, $T$ is the threshold and $I(x)$ is a function the result is one when $x$ is true and zero vice versa. The soft-thresholding function (also called the shrinkage function) takes the argument and shrinks it toward zero by the threshold. It is described as:

$$\eta_2(w) = (w - \text{sgn}(w)T)I(|w| > T), \quad (9)$$

where $\text{sgn}(x)$ is the sign of $x$. The soft-thresholding rule is chosen over hard-thresholding, the soft-thresholding method yields more visually pleasant images over hard-thresholding [7]. The BayesShrink function [8] has been attracting recently as an algorithm for setting different thresholds for every subband. Here subbands are frequently bands that differ from each other in level and direction. The BayesShrink function is effective for images including Gaussian noise. The observation model is expressed as follows: $Y = X + V$.

Here $Y$ is the wavelet transform of the degraded image, $X$ is the wavelet transform of the original image, and $V$ denotes the wavelet transform of the noise components following the Gaussian distribution $N(0, \sigma^2)$. Here, since $X$ and $V$ are mutually independent, the variances $\sigma_y^2$, $\sigma_x^2$ and $\sigma_v^2$ of $y$, $x$ and $v$ are given by:

$$\sigma_y^2 = \sigma_x^2 + \sigma_v^2. \quad (10)$$

Let us present a method for deriving of the noise: It has been shown that the noise standard deviation $\sigma_v$ can be accurately estimated from the first decomposition level diagonal subband $HH_1$ by the robust and accurate median estimator.

$$\hat{\sigma}_v^2 = \frac{median(|HH_1|)}{0.6745}. \qquad (11)$$

The variance of the degraded image can be estimated as

$$\hat{\sigma}_y^2 = \frac{1}{M}\sum_{m=1}^{M}A_m^2, \qquad (12)$$

where $A_m$ are the coefficients of wavelet in every scale $M$ is the total number of coefficient of wavelet. The threshold value $T$ can be calculated using

$$T_{MBS} = \frac{\beta\hat{\sigma}_v^2}{\hat{\sigma}_x}, \qquad (13)$$

where $\beta = \sqrt{\dfrac{\log M}{2 \times j}}$, $M$ is the total of coefficients of wavelet, $j$ is the wavelet decomposition level present in the subband coefficients under scrutiny and $\hat{\sigma}_x = \sqrt{\max(\hat{\sigma}_y^2 - \hat{\sigma}_v^2)}$. Note that in the case where $\hat{\sigma}_v^2 \geq \hat{\sigma}_y^2$, $\hat{\sigma}_x^2$ is taken to be zero, i.e. $T_{MBS} \to \infty$. Alternatively, in practice one may choose $T_{MBS} = \max|A_m|$, and all coefficients are set to zero.

In summary, the Modified BayesShrink thresholding technique performs soft thresholding with adaptive data driven subband and level dependent near optimal threshold given by:

$$T_{MBS} = \begin{cases} \dfrac{\beta\hat{\sigma}_v^2}{\hat{\sigma}_x}, & if\ \hat{\sigma}_v^2 < \hat{\sigma}_y^2 \\[2mm] \max|A_m|, & otherwise \end{cases} \qquad (14)$$

**Bilateral Filtering**
Bilateral filtering smooth the images while preserving edges, by means of a nonlinear combination of nearby image values [15]. The idea underlying bilateral filtering is to do in the range of an image what traditional filters do in its domain. Two pixels can be close to one another, that is, occupy nearby spatial location, or they can be similar to one another, that is, have nearby values, possibly in a perceptually meaningful fashion. Closeness refers to vicinity in the domain, similarity to vicinity in the range. Traditional filtering is a domain filtering, and enforces closeness by weighing pixel values with coefficients that fall off with distance. The range filtering, this averages image values with weights that decay with dissimilarity. Range filters are nonlinear because their weights depend on image intensity or color. Computationally, they are no more complex than standard non-separable filters. The combination of both domain and range filtering is termed as bilateral filtering. A low-pass domain filter to an image $f(x)$ produces an output image defined as follows:

$$\mathbf{h}(x) = k_d^{-1}(x)\int_{-\infty}^{\infty}\int_{-\infty}^{\infty}\mathbf{f}(\xi)\ c(\xi, x)\ d\xi, \qquad (15)$$

where $c(\xi, \mathrm{x})$ measures the geometric closeness between the neighborhood center $\mathrm{x}$ and a nearby point $\xi$. The bold font for $\mathbf{f}$ and $\mathbf{h}$ emphasizes the fact that both input and output images may be multiband. If low-pass filtering is to preserve the dc component of low-pass signals we obtain

$$k_d(x) = \int_{-\infty}^{\infty}\int_{-\infty}^{\infty}c(\xi, x)\ d\xi, \qquad (16)$$

If the filter is shift-invariant, $c(\xi, \mathrm{x})$ is only a function of the vector difference $\xi - \mathrm{x}$, and $k_d$ is constant. Range filtering is similarly defined:

$$\mathbf{h}(\mathbf{x}) = k_r^{-1}(\mathbf{x})\int_{-\infty}^{\infty}\int_{-\infty}^{\infty}\mathbf{f}(\xi)s(\mathbf{f}(\xi), \mathbf{f}(\mathbf{x}))d\xi. \qquad (17)$$

Except that now $s(\mathbf{f}(\xi), \mathbf{f}(\mathbf{x}))$ measures the photometric similarity between the pixel at the neighborhood center $\mathrm{x}$ and that of a nearby point $\xi$. Thus, the similarity function $s$ operates in the range of the image function $\mathbf{f}$, while the closeness function $c$ operates in the domain of $\mathbf{f}$. The normalization constant in Eq. (16) is replaced by

$$k_r(x) = \int_{-\infty}^{\infty}\int_{-\infty}^{\infty}s(\mathbf{f}(\xi), \mathbf{f}(\mathbf{x}))\ d\xi. \qquad (18)$$

Contrary to what occurs with the closeness function $c$, the normalization for the similarity function $s$ depends on the image $\mathbf{f}$. The similarity function $s$ is unbiased if it depends only on the difference $\mathbf{f}(\xi) - \mathbf{f}(\mathbf{x})$. The combined domain and range filtering will be denoted as bilateral filtering, which enforces both geometric and photometric locality. Combined filtering can be described as follows:

$$\mathbf{h}(\mathbf{x}) = k^{-1}(\mathbf{x})\int_{-\infty}^{\infty}\int_{-\infty}^{\infty}\mathbf{f}(\xi)c(\xi, \mathbf{x})s(\mathbf{f}(\xi), \mathbf{f}(\mathbf{x}))d\xi. \qquad (19)$$

with the normalization

$$k\ (\mathbf{x}) = \int_{-\infty}^{\infty}\int_{-\infty}^{\infty}c(\xi, \mathbf{x})s(\mathbf{f}(\xi), \mathbf{f}(\mathbf{x}))d\xi. \qquad (20)$$

**Contrast Stretching and Color Correction**
Contrast stretching often called normalization is a simple image enhancement technique that attempts to improve the contrast in an image by 'stretching' the range of intensity values. The full range of pixel values that the image concerned is given by Eq. (21). Color correction is performed by equalizing each color means. In underwater image colors are rarely balanced correctly, this processing step suppresses prominent blue or green color without taking into account absorption phenomena.

$$I_{i,j} = \begin{cases} \dfrac{I_{i,j} - \min_I}{\max_I - \min_I} & if \quad 0 < I_{i,j} < 1 \\ 0 & if \quad 0 > I_{i,j} \\ 1 & if \quad 1 < I_{i,j} \end{cases} \qquad (21)$$

where $\min_I$ and $\max_I$ are the minimum and maximum intensity values in the image.

## 3. EXPERIMENTAL RESULTS

We have conducted experiments to evaluate our preprocessing technique on degraded underwater images with unknown turbidity characteristics. The scene includes several objects at distance [1m, 2m], near the corner of the water body. The images were captured using Canon D10 water proof camera at a depth of 2m from the surface level of water. The captured images are diminished due to optical properties of the light in an underwater environment. These images suffer from non-uniform illumination of light, low contrast, blurring and typical noise levels for underwater conditions. The preprocessing technique comprises of homomorphic filtering to correct non-uniform illumination of light, wavelet denoising to remove additive Gaussian noise present in underwater images, bilateral filtering to smooth underwater image and contrast stretching to normalize the RGB values. The experimentation is carried out in two stages. In the first stage, we have conducted various experiments on captured images and estimated optimal parameters for bilateral filter. Similarly, optimal filter bank and optimal wavelet shrinkage function are estimated for wavelet denoising. In the second stage, we conducted the experiments using estimated optimal parameters, optimal filter bank and optimal wavelet shrinkage function for evaluating the proposed technique.

The procedure involved in the first stage is as follows: after applying the homomorphic filter for correction of illumination and reflectance components, wavelet denoising is used to remove the Gaussian noise, which is common in the underwater environment. In wavelet denoising, filter bank plays an important role for the best result of denoising. We performed an evaluation of four filter banks such as Haar, db4, Sym4 and Coif4 for decomposing the image prior to applying shrinkage function. The Table I shows the PSNRs obtained using four filter banks such as Haar, db4, Sym4 and Coif4 for each underwater image. The Coif4 filter bank yields optimal PSNR for all the underwater images.

After finding the Coif4 filter bank is the best for underwater images, we identify suitable wavelet shrinkage function by comparing and evaluating various wavelet shrinkage functions based on PSNR. We have considered BayesShrink, VisuShrink, Adaptive Subband Thresholding, NormalShrink and Modified BayesShrink for comparison purpose [8-14]. The Table II presents comparison results of various wavelet shrinkage functions. The experimental result shows that the Modified BayesShrink function achieves highest PSNR compared to other shrinkage functions. Hence, the wavelet based denoising technique with Modified BayesShrink function is suitable for removing the additive noise present in the underwater images. The Modified BayesShrink performs denoising that is consistent with the human visual system that is less sensitive to the presence of noise in a vicinity of image features.

We applied bilateral filter on denoised image for edge preserving smoothing that is non-iterative and simple. The bilateral filtering is performed in CIE Lab color space is the natural type of filtering for color images. To find the optimal parameter for bilateral filter, the bilateral filtering is applied to denoised images by varying the parameters $\sigma_d$ and $\sigma_r$. The interval of parameter values considered, for $\sigma_d$ is 1-10 pixel values and for $\sigma_r$ is 10-200 intensity values. The experimental results show that the parameter values $\sigma_d$ =1 and $\sigma_r$ =10 smooth the image compared to other parameter values. Hence, $\sigma_d$ =1 and $\sigma_r$ =10 are the optimal parameters for bilateral filter.

In the second stage of experimentation, we carried out various experiments on captured images to evaluate the proposed technique. In the first stage, we identified that $\sigma_d$ =1 and $\sigma_r$ =10 for bilateral filter, similarly, Coif4 filter bank and Modified BayesShrinkage function for wavelet denoising yields best results on degraded underwater images. We used the same setup to evaluate the technique quantitatively as well as qualitatively. In order to evaluate quantitatively, the gradient magnitude histogram for original and preprocessed images are estimated and shown in "Fig. (3)". These histograms show that gradient values are larger after preprocessing compared to gradient values obtained on original images. For the qualitative evaluation, the edge detection is applied on both original and preprocessed images, shown in "Fig. (2)". The edge detection results of the preprocessed images demonstrate an efficiency of the proposed technique.

To verify the processing time of bilateral filter with anisotropic filter, we carried out an experiments using the same setup on degraded images. This shows the tendency of the bilateral filter's time cost. The results were obtained on an Intel Pentium Core i5 processor of speed 3.30 GHz and 4 GB of RAM. The techniques were implemented and tested in MATLAB environment. The proposed technique with bilateral filter spends 11.238591 seconds where as the proposed technique with anisotropic filter spends 14.443145 seconds. The proposed technique with bilateral filter requires less computation time; the important speed gain is due to the use of non-iterative procedure, which drastically reduces the number of operations.

## 4. CONCLUSION

In this paper, we proposed a preprocessing technique for enhancing the quality of degraded underwater images. The proposed technique includes four filters such as homomorphic filtering, wavelet denoising, bilateral filtering and contrast equalization, which are applied sequentially. The main contribution of this paper is inclusion of bilateral filter for smoothing in addition to existing other filtering techniques. We identified that $\sigma_d$ =1 and $\sigma_r$ =10 for bilateral filter, similarly, combination of Coif4 filter bank and Modified BayesShrink function yields higher PSNR values. The processing time of the proposed technique is very low compared to the preprocessing technique with anisotropic filtering. The proposed preprocessing technique enhances the quality of the degraded underwater images which are suffered from non-uniform illumination, low contrast, noise and diminished colors. The quantitative and qualitative evaluation results demonstrate improved performance of the technique.

In order to perform a quantitative comparison of results with other standard preprocessing algorithms for underwater images, a common database should be available to test according to specific criteria. To our knowledge, no such underwater databases exist at present. Therefore, the proposed technique results are not compared with other standard preprocessing algorithms. The development of underwater image database could be one of the future research lines from which the underwater community would certainly beneficiate.

## ACKNOWLEDGEMENT

The authors are grateful to the reviewers for their critical comments and numerous suggestions. The research work was supported by Naval Research Board (Grant No.158/SC/08-09), DRDO, New Delhi, India.

*Table - 1- The comparison of four wavelet filter banks based on PSNR (dB):*

| Image # | Image | Filter Bank | MSE | PSNR (dB) |
|---|---|---|---|---|
| **Image #1** | 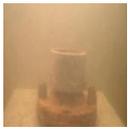 | Haar | 0.0242 | 64.2905 |
| | | Db4 | 0.0050 | 71.1837 |
| | | Sym4 | 0.0059 | 70.4499 |
| | | **Coif4** | **0.0043** | **71.7861** |
| **Image #2** | 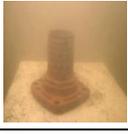 | Haar | 0.0540 | 60.8066 |
| | | Db4 | 0.0144 | 66.5616 |
| | | Sym4 | 0.0161 | 66.0607 |
| | | **Coif4** | **0.0128** | **67.0677** |
| **Image #3** | 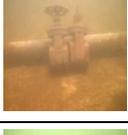 | Haar | 0.1909 | 55.3232 |
| | | Db4 | 0.2131 | 54.8456 |
| | | Sym4 | 0.1217 | 57.2787 |
| | | **Coif4** | **0.1154** | **57.5078** |
| **Image #4** | 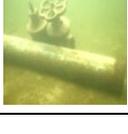 | Haar | 0.0744 | 59.4173 |
| | | Db4 | 0.1129 | 57.6021 |
| | | Sym4 | 0.0643 | 60.0509 |
| | | **Coif4** | **0.0594** | **60.3960** |

*Table - 2 - The comparison of five wavelet shrinkage functions based on PSNR (dB):*

| Image # | Modified BayesShrink | BayesShrink | NormalShrink | Adaptive Subband Thresholding | VisuShrink |
|---|---|---|---|---|---|
| **Image #1** | **66.2116** | 66.2008 | 65.9764 | 65.0434 | 50.6137 |
| **Image #2** | **63.1458** | 63.1301 | 62.8485 | 61.6751 | 48.3258 |
| **Image #3** | **54.8456** | 54.7995 | 54.0704 | 52.1615 | 41.7070 |
| **Image #4** | **57.6021** | 57.5921 | 57.4179 | 56.6197 | 44.7697 |

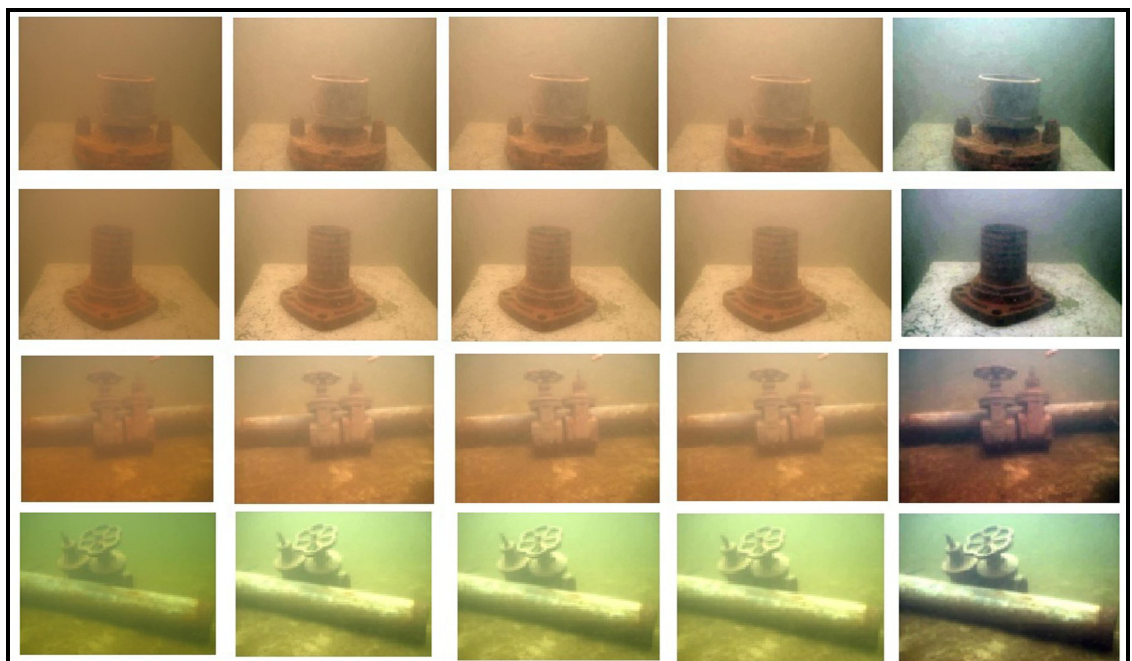

Fig. 1: First column: original image, second column: after homomorphic filtering, third column: after wavelet denoising, fourth column: after bilateral filtering, last column: after contrast equalization

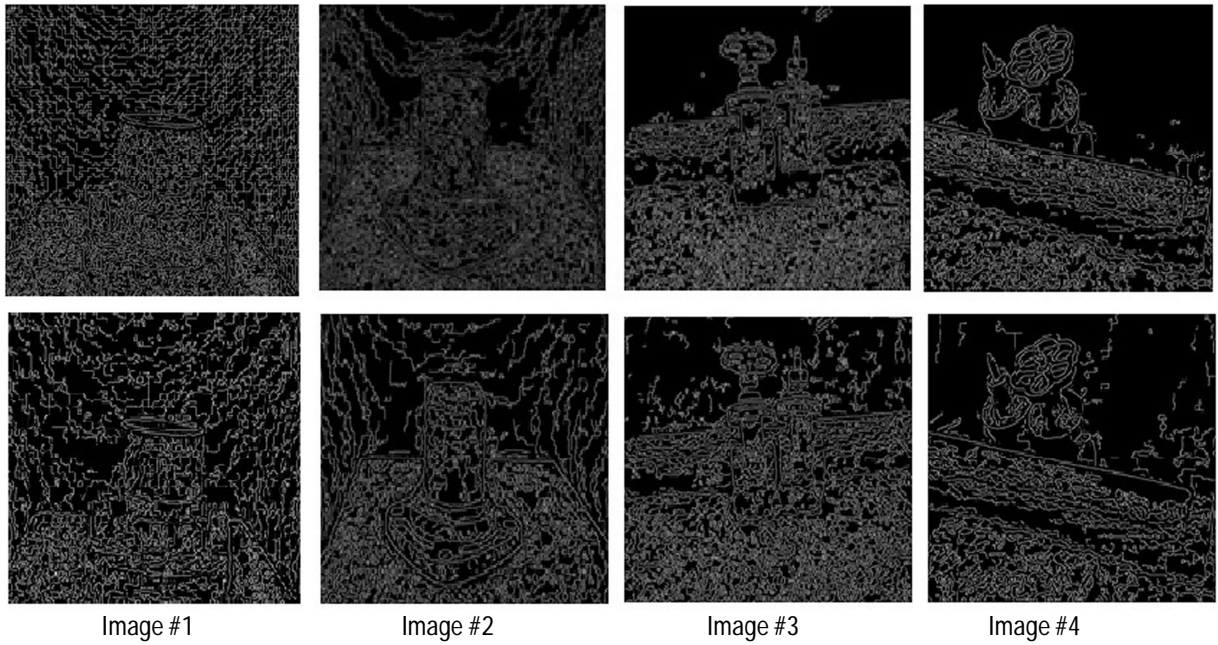

Image #1      Image #2      Image #3      Image #4

Fig. 2: Edge detection results on four images;
First row: edge detection results on original images, Second row: edge detection results on preprocessed images

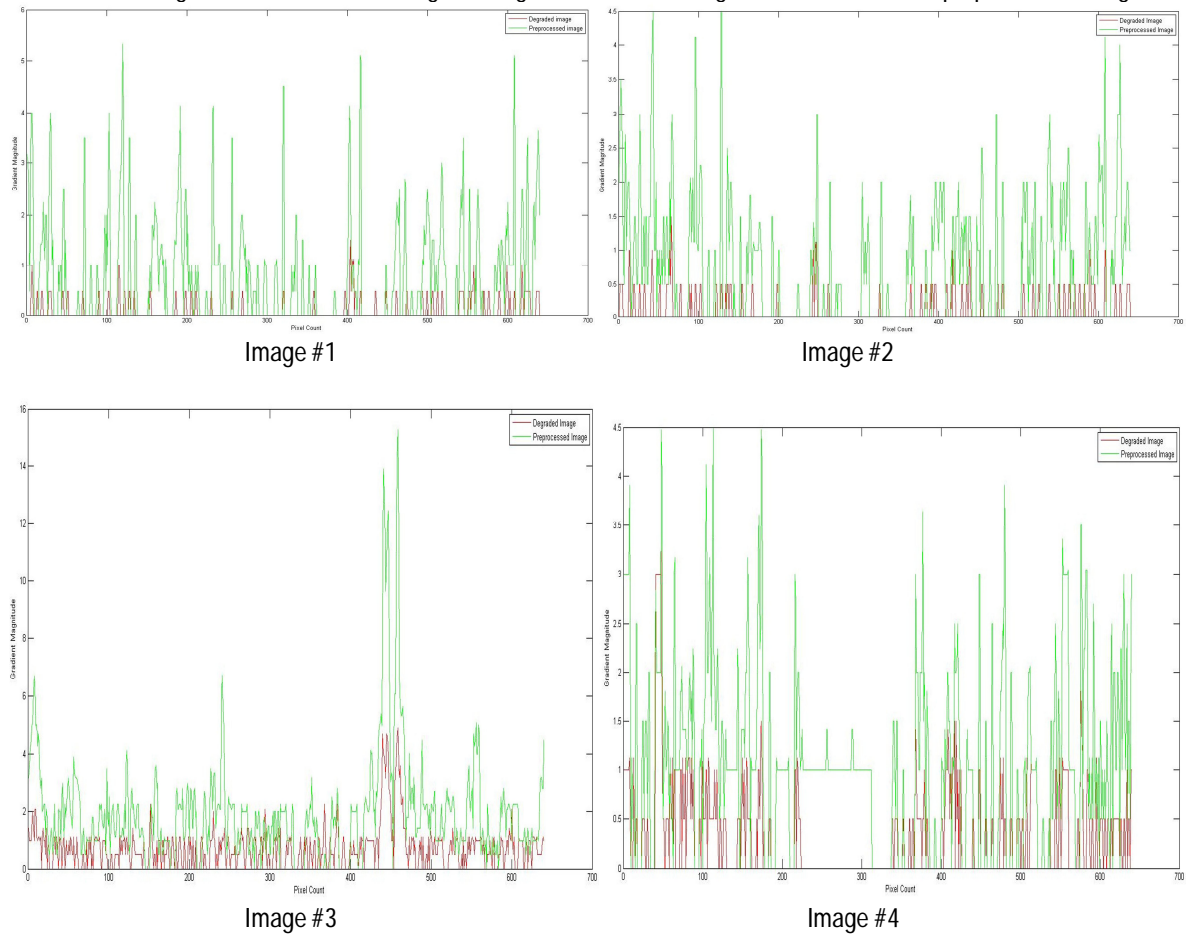

Image #1      Image #2

Image #3      Image #4

Fig. 3: Gradient magnitude histogram of the four images;
Red line: the gradient magnitude histogram of the original image, Green line: the gradient magnitude histogram of the preprocessed image